\documentclass{trbunofficial}
\usepackage{graphicx}
\usepackage{amsmath}
\usepackage{amssymb}
\usepackage{array}
\usepackage{booktabs}
\usepackage{pifont}

\AuthorHeaders{Gan, Shi, P.Li, An, L.Li, J.Ma, C.Ma, Ran}
\title{Goal-based Neural Physics Vehicle Trajectory Prediction Model}

\author{%
  \textbf{Rui Gan}\\
  University of Wisconsin-Madison, Madison, WI, USA\\
  Email: rgan6@wisc.edu\\
  \hfill\break
  \textbf{Haotian Shi, Ph.D., Corresponding Author}\\
  University of Wisconsin-Madison, Madison, WI, USA\\
  Email: hshi84@wisc.edu\\
  \hfill\break
  \textbf{Pei Li, Ph.D., Corresponding Author}\\
  University of Wisconsin-Madison, Madison, WI, USA\\
  Email: pei.li@wisc.edu\\
  \hfill\break
  \textbf{Keshu Wu, Ph.D.}\\
  University of Wisconsin-Madison, Madison, WI, USA\\
  \hfill\break
  \textbf{Bocheng An}\\
  Southeast University, Nanjing, Jiangsu, China\\
  \hfill\break
  \textbf{Linheng Li, Ph.D.}\\
  Southeast University, Nanjing, Jiangsu, China\\
  \hfill\break
  \textbf{Junyi Ma}\\
  University of Wisconsin-Madison, Madison, WI, USA\\
  \hfill\break
  \textbf{Chengyuan Ma, Ph.D.}\\
  University of Wisconsin-Madison, Madison, WI, USA\\
  \hfill\break
  \textbf{Bin Ran, Ph.D.}\\
  University of Wisconsin-Madison, Madison, WI, USA\\
  \hfill\break
}

\begin{document}
\maketitle

\section{Abstract}

Vehicle trajectory prediction plays a vital role in intelligent transportation systems and autonomous driving, as it significantly affects vehicle behavior planning and control, thereby influencing traffic safety and efficiency. Numerous studies have been conducted to predict short-term vehicle trajectories in the immediate future. However, long-term trajectory prediction remains a major challenge due to accumulated errors and uncertainties. Additionally, balancing accuracy with interpretability in the prediction is another challenging issue in predicting vehicle trajectory. To address these challenges, this paper proposes a \textbf{G}oal-based \textbf{N}eural \textbf{P}hysics Vehicle Trajectory Prediction Model (\textbf{GNP}). The GNP model simplifies vehicle trajectory prediction into a two-stage process: determining the vehicle's goal and then choosing the appropriate trajectory to reach this goal. The GNP model contains two sub-modules to achieve this process. The first sub-module employs a multi-head attention mechanism to accurately predict \textbf{goals}. The second sub-module integrates a deep learning model with a physics-based social force model to progressively predict the complete trajectory using the generated goals. The GNP demonstrates state-of-the-art long-term prediction accuracy compared to four baseline models. We provide interpretable visualization results to highlight the multi-modality and inherent nature of our neural physics framework. Additionally, ablation studies are performed to validate the effectiveness of our key designs.

\hfill\break%
\noindent\textit{Keywords}: Vehicle Trajectory prediction, Neural differential equations, Goal-based Prediction, Transformer, Vehicle Intentions
\newpage

\section{1. Introduction}
Vehicle trajectory prediction is a crucial component in intelligent transportation systems (ITS) and autonomous driving, with significant implications for vehicle behavior planning and control. On one hand, the collection of high-quality vehicle trajectory data is becoming more feasible as vehicles can promptly share accurate position and speed information, along with data about surrounding vehicles and road environments gathered through various sensors, using high-quality communication networks like 5G. On the other hand, for autonomous vehicles (AVs) and connected autonomous vehicles (CAVs), precise trajectory prediction is a necessary prerequisite for reliable decision-making, trajectory planning, and control instructions \cite{shi2023deep}. Effectively leveraging vehicle trajectory data, as well as information about surrounding vehicles and road environments, to achieve high-performance prediction is an imperative task in ITS and autonomous driving.

Earlier models for vehicle trajectory prediction relied on physical models or traditional machine learning algorithms. Physics-based models employ mechanics or kinematics equations, iterating the prediction step to forecast the trajectory \cite{lefevre_survey_2014}. Traditional machine learning methods included Bayesian learning, hidden Markov models (HMMs), support vector machines (SVMs), and Gaussian Processes (GP) \cite{goli2018vehicle}. However, these methods often struggled with long-term predictions as they could not anticipate changes due to maneuvers or external factors and required extensive feature engineering. More recently, deep learning models have revolutionized vehicle trajectory prediction and perform great accuracy by demonstrating the ability to extract intricate temporal and spatial features from data. Recurrent neural networks (RNNs), including long short-term memory network (LSTM) and gated recurrent unit (GRU) variants \cite{altche2017lstm, benterki2019long, park2018sequence}, and transformer-based models \cite{chen2022vehicle, zhang2022trajectory} excel in capturing temporal features from sequential trajectory data. For spatial features and interactions, convolutional neural networks (CNNs) \cite{deo_convolutional_2018, messaoud2019non, wang2020multi}, and graph neural networks (GNNs) \cite{wu_graph-based_2023, li2019grip} are employed to analyze Euclidean or non-Euclidean spatial dependencies between vehicles and the infrastructures. Additionally, to model the high uncertainty of vehicle trajectories, some models adopt a multi-modality approach to predict multiple possible motion behaviors. These models include those using predefined explicit driving maneuvers or pre-clustered anchors, as well as generative deep learning models such as variational autoencoders (VAEs) \cite{feng2019vehicle} and generative adversarial networks (GANs) \cite{wang2020multi, gupta2018social}. However, despite their high accuracy, deep learning models have limitations. They require large amounts of labeled data, are computationally intensive, and often act as "black boxes" with limited interpretability, which may hinder safety-critical applications in autonomous driving.

Although previous trajectory prediction studies have achieved notable success, we have identified and analyzed two key issues from these findings. First, vehicle trajectory prediction has uncertainties, largely due to the unknown intention of the vehicle. The uncertainty in a vehicle's trajectory primarily arises from two main factors: the unknown driving intention of the driver over a specific period and the indeterminate path the vehicle will take to fulfill this intention. In a highway environment, the former has a significant influence on trajectory prediction outcomes. For instance, whether a vehicle chooses to proceed straight or change lanes leads to completely different trajectory patterns. Additionally, the behavior of surrounding vehicles—such as slowing down, speeding up, and lane changes—directly influences the current vehicle's decisions. The latter is also relevant, as once the vehicle determines its intention (goal), there may be several path options to execute. Previous studies primarily inferred future motion states from historical trajectories and environmental information, i.e., focusing solely on the latter factor but neglecting the modeling of driving intentions. Consequently, designing a model that can accurately understand and distinguish a vehicle's intention is crucial for precise trajectory prediction. Recently, goal-based prediction frameworks have gained popularity and proven effective in trajectory prediction tasks \cite{ghoul_lightweight_2022, tran_goal-driven_2021, ghoul_interpretable_2023, zhao2021tnt, gu_densetnt_2021, mangalam_it_2020, avidan_human_2022}. In terms of vehicle trajectories, studies such as \cite{zhao2021tnt} and \cite{gu_densetnt_2021} feed historical trajectories and context information into neural network layers to explicitly output grid-based goals and then complete the full trajectories. Nonetheless, the goal prediction module in these models relies solely on the historical trajectory segment, neglecting the future general pattern of vehicle trajectories. This oversight makes it challenging to accurately predict the distribution of possible goals.

Secondly, previous models often face a dilemma between interpretability and data-fitting ability. These models can be broadly categorized into two types: physics-based and data-driven models. Physics-based models offer good interpretability as they are constructed using explicit geometric optimization or ordinary/partial differential equations. They typically use combinations of physical quantities, formulas, and parameters to obtain results. However, the parsimonious structures of a physics model may not always effectively capture the complex nonlinearities, high dimensionality, or latent patterns of vehicular trajectory data, thus limiting its predictive capability. In contrast, data-driven models excel at fitting data because they can be trained on large datasets with numerous parameters. However, their black-box nature hinders researchers from interpreting the predictions. Relying solely on such models without understanding their internal working mechanisms poses safety risks. Interpretability allows researchers to identify potential errors and avoid risks. Moreover, interpretability provides transparency, enabling users and decision-makers to trust the model's output. Therefore, developing a vehicle trajectory prediction model that balances prediction accuracy and interpretability is critical. \cite{geng_physics-informed_2023} proposed a PIT-IDM model that integrates the Transformer with a car-following model, the Intelligent Driver Model (IDM). By employing the novel physics-informed neural networks (PINNs) framework, this model achieves great prediction accuracy and transferability and delivers highly explainable visualizations. However, they train the deep-learning and physics components separately and merge them only in the final loss function, thereby not fully exploiting their combined potential benefits. Additionally, their model only investigates vehicle trajectory prediction in the longitudinal direction.

To tackle the aforementioned issues, we propose a model that predicts the intentions of vehicles while also ensuring interpretability. We construct a model with two sub-modules: a goal prediction sub-module and a trajectory prediction sub-module. The model first estimates possible future goals based on historical data, and these goals guide the trajectory. Next, we combine a physical model with neural networks for trajectory completion in the latter sub-module. Drawing inspiration from recent advancements in neural differential equations \cite{huang2011social, avidan_human_2022}, we utilized a deterministic social force model. Unlike traditional social force models and their variants, we replace fixed or calibrated parameters with learnable neural networks. This method utilizes the superior data-fitting abilities of neural networks to more accurately represent the dynamics of variables in the physical model. Overall, the first sub-module effectively exploits complex uncertainty by estimating dynamic vehicle motion goals, while the second sub-module leverages the benefits of both physics-based models and deep learning to achieve good data-fitting and interpretability.

The primary contributions of this paper are as follows:

\rightskip=2em 

\begin{enumerate}
    \item We explore the general pattern of vehicle trajectories to investigate possible driving intentions, explicitly predicting multiple potential goals over a period of time.
    \item We design a neural differentiable equation model to forecast the complete trajectory, calculating the key parameters of attraction and repulsion forces in the Social Force Model through neural networks and progressively determining future positional coordinates based on vehicle dynamics.
    \item  We introduce a novel Goal-based Neural Physics Vehicle Trajectory Prediction Model (GNP) that integrates deep learning models with physical social force models to achieve both high prediction accuracy and interpretability. We demonstrate the effectiveness of this model through extensive experiments and visualizations.
\end{enumerate}

\rightskip=0em 

\section{2. Methodology}
In this study, we focus on predicting the trajectory of the target vehicle in a multi-lane highway scenario based on the historical movement states of the target vehicle and its neighborhood vehicles. Our methodology emphasizes goal-based prediction, considering the interaction between the target vehicle and its surrounding vehicles and road marking information. 

\subsection{2.1 Problem Definition}
We suppose that a sequence of vehicle state from time $t=1$ to $t=T$ as $Q_{1:T} = (q^1, q^2, \ldots, q^T)$, where an observation or state of the vehicle at time $t$ is denoted as $q^t = \begin{bmatrix} p^t, \dot{p}^t \end{bmatrix}^\text{T}$, $p, \dot{p} \in \mathbb{R}^2$ denotes the position and velocity. For a single state $q_i^t$ of vehicle $i$, we also consider its neighborhood set $\Omega_i^t \in \mathbb{R}^N$ comprising its surrounding vehicle states $\{ q_j^t : j \in \Omega_i^t \}$. Therefore, the observed sequence of vehicle state is $X = Q_{1:T_{obs}}$ and the future or ground truth sequence is $Y = Q_{T_{obs}+1:T}$. The predicted trajectory is denoted as  $\hat{Y} = (\hat{p}^{T_{obs}+1}, \hat{p}^{T_{obs}+2}, \ldots, \hat{p}^T)$. We aim to predict the position trajectory of the target vehicle.

The vehicle driving process for a specific period is summarized as follows. Firstly, within the time scope of a trajectory $T$, the drivers decide on a set of possible destinations on the highway, which we call goals $g \in \{1, 2, \ldots, n_{\text{goals}}\}$, where $n_{\text{goals}}$ denotes the number of goals; second, oriented by the potential goals and considering the influence of surrounding vehicles and road markings, they choose one goal and manipulate the vehicle towards it via a route. In a 2D scene, the goal $g$ of the target vehicle is defined as the last longitudinal and lateral position of the future time step, denoted as $H_{T+F}$. Therefore, the goal-based prediction model can be presented as follows:

\begin{linenomath}
  \begin{flalign}
    & \hat{g} = G_{\mu}(X, \Omega), \\
    & \hat{Y} = F_{\theta, \phi}(X, \Omega, \hat{g}, E),
  \end{flalign}
\end{linenomath}

\begin{linenomath*}
\vspace{1em}
\end{linenomath*}

\noindent where $G$ and $\mu$ are the Goal Prediction sub-module in \textbf{Section 2.3} and its learnable parameters (neural network weights), $\hat{g}$ is the estimated goals of the Goal Prediction sub-module. Similarly, in \textbf{Equation 2}, $F$ is the Goal-based Neural Social Force Trajectory Prediction sub-module introduced in \textbf{Section 2.4}, $\theta$ and $\phi$ are interpretable parameters presented later and uninterpretable parameters in neural networks. Here, $E$ represents the environmental information.

\begin{figure}[!ht]
  \centering
  \includegraphics[width=1\textwidth]{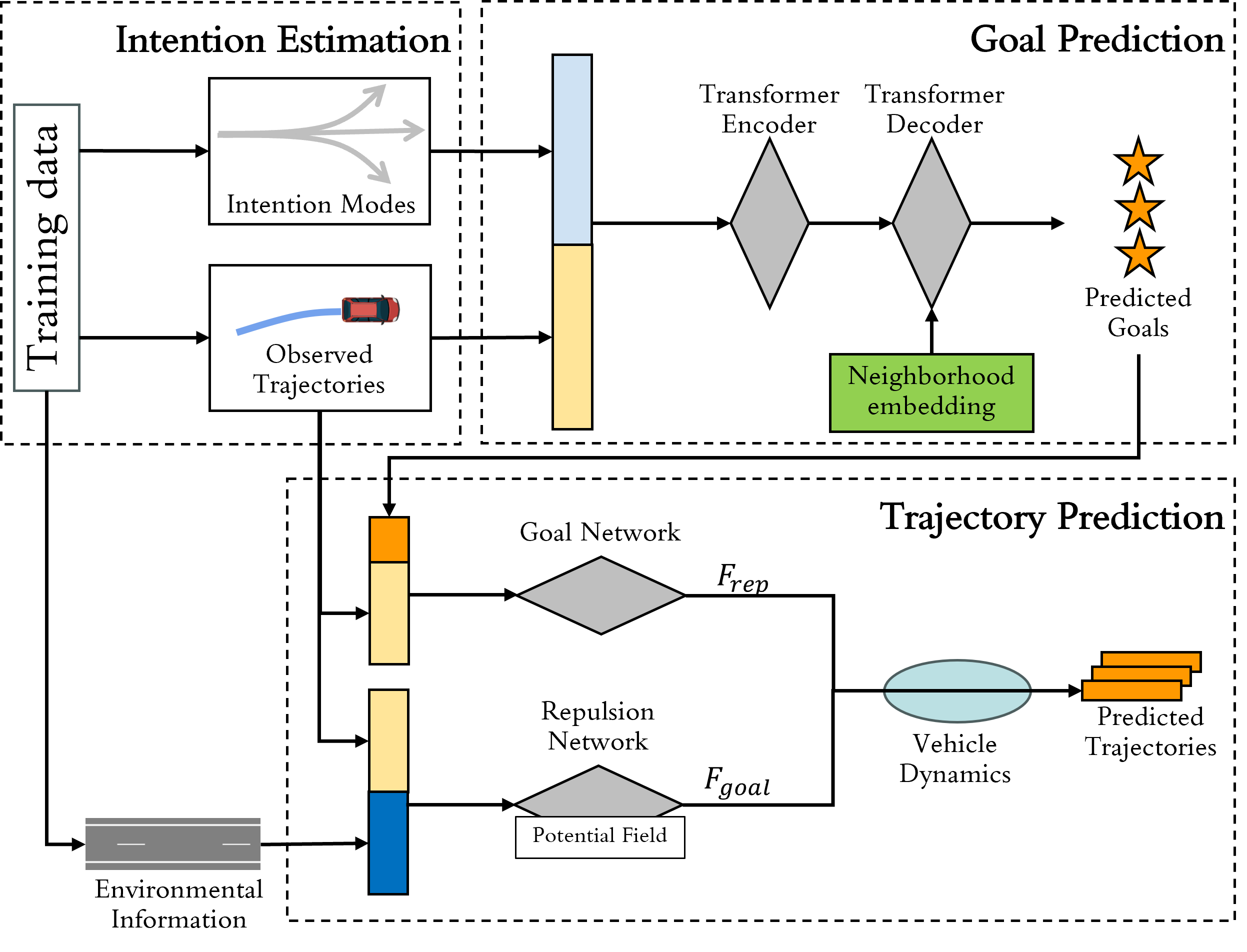}
  \caption{Model architecture proposed in this paper. This dual sub-module framework first estimates the intentions and predicts multiple possible goals, then forecasts the full trajectories using a deep-learning enhanced social force model.}\label{fig framework}
\end{figure}

\subsection{2.2 Model Architecture}
Based on the defined problem, we introduce the architecture of our GNP model in this section. As depicted in Figure~\ref{fig framework}, the GNP model comprises two sub-modules: the Goal Prediction sub-module and the Trajectory Prediction sub-module. The Goal Prediction sub-module employs a refined transformer encoder-decoder architecture to generate diverse goals for vehicles. This sub-module encodes observed historical trajectories and a future trajectory trend (intention mode) obtained through clustering. It decodes the observed trajectories of neighboring vehicles to capture social interactive dependencies. By parsing the vehicle's historical trajectories, surrounding vehicle information, and intention modes, the Goal Prediction sub-module models the vehicle's intentions and samples multiple goals for the subsequent trajectory prediction.

Following this, the Trajectory Prediction sub-module employs a new neural differential equations model to predict the complete trajectory by computing acceleration values in conjunction with the vehicle dynamics. Specifically, we achieve this model by integrating a physical model, the social force, with a neural network. To accurately evaluate the vehicle's acceleration, the Trajectory Prediction sub-module uses previously generated goals and observed trajectories to train and learn the key parameters of the two components of the social force. Finally, full future trajectories are predicted step by step by adhering to the aforementioned vehicle dynamics.

\subsection{2.3 Goal Prediction}
To model intentions and predict goals for vehicles, we developed an enhanced transformer encoder-decoder architecture based on the design in \cite{shi_trajectory_2023}. First, the mode-level transformer encoder parses the intention modes and observed embeddings, capturing possible future trends in vehicle trajectories. The encoder's output, combined with information about neighboring vehicles, is fed into the social interaction-level transformer decoder to explore inter-vehicle relationships. Finally, the model produces multiple possible goals, determining the vehicle's driving intention over a specified period.

In contrast to other transformer-based trajectory prediction models, our model aims to predict goals by estimating driving intentions. Previous methods typically depend only on observed trajectories and neighboring vehicles, encoding temporal and spatial features from these sequences. Our model, however, incorporates general patterns of future vehicle behavior into the input token, enabling it to identify potential future driving directions and targets and thereby predict the goals.

\vspace{1em}
\noindent \textbf{Trajectory modes and observed embeddings:} The input token of our model combines the designed intention mode and the observed trajectory sequence. By interpreting this token with a mode-level transformer encoder, the model captures the overall pattern and trend of the trajectory sequence. The intention mode differentiates general driving behaviors on the highway, indicating various driving purposes. For example, lane-changing trajectories differ significantly from straight-ahead trajectories, allowing us to infer whether a vehicle intends to change lanes or go straight. However, determining intention solely based on subjective evaluations like going straight or changing lanes is insufficient. Therefore, we first perform two rigid transformations—translation and rotation—on the input trajectory sequence and then obtain the intention mode using a distance-based method.

From a bird's eye view of the highway, the trajectory segments are distributed across various parts of the roadway and travel in two different directions. To extract the general features of the vehicle trajectories, it is essential to normalize the trajectories using two rigid transformations: translation and rotation. It is important to note that trajectories exhibit rigid transformation invariance. For instance, the trajectory of a vehicle moving from east to west and changing lanes to the right can be rotated and translated to represent the same vehicle moving from west to east and changing lanes to the right under identical conditions. The trajectory data in the training set is divided into two segments: the first segment with a length of $T_{obs}$ and the second segment with a length of $T_{pred}$. First, we translate the entire trajectory by shifting the $T_{obs}+1$ point of each trajectory to the origin of the coordinate system, i.e.,  subtracting the 2D coordinates of the $T_{obs}+1$ point from all trajectory points. Then, considering the presence of trajectories with two travel directions on the highway in some datasets, we rotate all trajectories around the coordinate origin and uniformly convert them to trajectories traveling from west to east. For the transformed trajectories, the distance between the trajectory points with similar intention behavior is smaller. Hence, a distance-based approach can be used here to distinguish the different intention modes of the trajectories.

Here, we perform a clustering operation on the normalized future trajectories to identify \(L\) centers \(C \in \mathbb{R}^{L \times T_{pred} \times 2}\), with \(C = \{c_1, \ldots, c_L\}\) and each \(\{c_l \mid l \in \{1, \ldots, L\}\}\) representing a trajectory of length \(T_{pred}\). These centers \(C\) represent the general intention modes and act as one of the inputs for the next mode-level transformer encoder. Staying consistent with the original text, the clustering operation here is a preparatory step, $C$ does not affect the efficiency of validation or test stage.

The intention mode requires two more processing steps to serve as the input token for the next mode-level transformer encoder. They are first reshaped into a $L \times 2T_{pred}$ features and then embedded by a learnable linear transformation, resulting in the input embedding $E_c$ as follows:

\begin{equation}
E_c = \phi(C, W_c),
\end{equation}

\begin{linenomath*}
\vspace{1em}
\end{linenomath*}

\noindent where $\phi(\cdot, \cdot)$ denotes a linear transformation characterized by a learnable parameter matrix $\mathbf{W_c} \in \mathbb{R}^{2T_{pred} \times D_e}$, and $E_c \in \mathbb{R}^{L \times D_e}$ denotes the intention mode embedding.

To ensure accurate goal prediction, it is necessary to analyze the observed historical trajectories in addition to the intention mode. Therefore, the intention mode and observed trajectory $X \in \mathbb{R}^{B \times T_{obs} \times 2}$ are combined here. Specifically, the embedded $X$ is concatenated to the intention mode embedding  $E_c$ as follows:

\begin{linenomath}
  \begin{flalign}
    & E_o = \phi(X, W_o), \\
    & E_e = E_c + E_o,
  \end{flalign}
\end{linenomath}

\begin{linenomath*}
\vspace{1em}
\end{linenomath*}

\noindent where $\mathbf{B}$ denotes the batch size, $\mathbf{X}$ denotes reshaped into $B \times 2T_{obs}$ followed by a linear transformation, $\mathbf{W_o} \in \mathbb{R}^{2T_{obs} \times D_e}$ denotes the learnable parameter matrix. The dimensions of $E_c$ and $E_o$ are broadcast to $B \times L \times D_e$, followed by an addition operation to obtain the input embedding $E_e \in \mathbb{R}^{B \times L \times D_e}$. Classical Transformer models typically require positional encoding to incorporate sequential order information into the sequence. However, unlike tasks such as natural language processing, the input data in vehicle trajectory prediction inherently possesses temporal sequential information. Therefore, following the approach in \cite{shi_trajectory_2023}, we do not add positional embedding to the input embedding.

\vspace{1em}
\noindent \textbf{Mode-level transformer encoder:} The mode-level transformer encoder is designed to analyze intention modes and observed trajectories to predict the vehicle's potential goals or destinations over time. Given the input embedding $E_e$, which represents general intention modes and the observed trajectory, the mode-level transformer encoder employs the standard encoder architecture of a naive transformer. Each encoder block features a multi-head self-attention layer and a Feed-Forward Network (FFN) with residual connections \cite{he2016deep}. Unlike the naive transformer encoder, we do not add positional embedding at each encoder block as proposed in \cite{shi_trajectory_2023}, but only provide the combined input embedding once.

\vspace{1em}
\noindent \textbf{Social interaction-level transformer decoder:} The social-level transformer decoder is designed to extract social interactions with neighboring entities and follows the standard decoder architecture of a naive transformer, which includes an attention layer and a Feed-Forward Network (FFN). Referring to the work of \cite{shi_trajectory_2023}, the differences from the naive transformer are threefold: 1. The decoder receives masked neighboring embeddings instead of masked output embeddings. The mask's main role is to exclude non-existent neighbors. 2. The decoder retains the encoder-decoder attention mechanism but omits self-attention. 3. Positional embeddings are removed since the positional relationship between the vehicle and its neighbors is incorporated in the trajectory sequences.

Assume that a pedestrian has $N$ neighbors, represented by the neighbor observed trajectories $X_s \in \mathbb{R}^{N \times T_{obs} \times 2}$. Each trajectory in $X_s$ is flattened into a feature vector, leading to a feature matrix $\hat{X}_s \in \mathbb{R}^{N \times 2T_{obs}}$. Then, we embed the feature matrix by a learnable linear transformation to obtain the input embeddings of the social-level transformer decoder as follows:

\begin{equation}
E_s = \phi(\hat{X}_s, W_s),
\end{equation}

\begin{linenomath*}
\vspace{1em}
\end{linenomath*}

\noindent where $E_s \in \mathbb{R}^{N \times D_e}$ is the input embeddings of the decoder, $W_s \in \mathbb{R}^{2T_{obs} \times D_e}$ is the learnable parameter matrix. After that, the input embeddings $E_s$ are transformed into output embeddings with the subsequent encoder-decoder attention layer and an FFN layer with the residual connection. In this case, these output embeddings attend to the social interactions to forecast social-acceptable trajectories and corresponding probabilities by the next dual prediction.

\vspace{1em}
\noindent \textbf{Output:} To accurately describe the possible driving intentions of the vehicle, we output multiple potential goals and their corresponding probabilities. Based on the literature \cite{shi_trajectory_2023}, we use a goal prediction head and a probability prediction head to output the predicted goals and their probabilities. Here, the probability prediction head is directly connected to the mode-level transformer encoder, while the goal prediction head is applied after the social interaction-level transformer decoder. We adopt different strategies during the training and deployment phases.  In training, we use a greedy strategy, assuming that the predicted goal with the highest probability originates from the cluster center nearest to the ground truth goal \(p^T\). Specifically, we first calculate the distance between the cluster centers \(C = \{c_1, \ldots, c_L\}\) and the ground truth goal \(p^T\) to identify the nearest clustering center \(c_i , i \in \{1, \ldots, L\}\). 

\begin{equation}
i = \arg\min_{i \in \{1, \ldots, L\}} \left( \| \hat{Y} - c_i \|_2^2 \right).
\end{equation}

\begin{linenomath*}
\vspace{1em}
\end{linenomath*}

Then, the soft probability \(\hat{p}\) of \(c_i\) can be expressed using the normalized negative distance as shown below.

\begin{equation}
p = \text{softmax}(\{ -\| \hat{Y} - c_i \|_2^2 \mid i \in \{1, \ldots, L\} \}).
\end{equation}

\begin{linenomath*}
\vspace{1em}
\end{linenomath*}

\noindent As such, the predicted goal \(\hat{p}^T\) and its corresponding probability \(\hat{p}\) are derived by applying a series of deep transformations to \(c_i\). In the testing phase, we adopt a Top-K strategy, i.e., we select the K highest probability goals from the model's multiple outputs to represent the final set of possible driving goals.

\subsection{2.4 Goal-based Neural Social Force Trajectory Prediction}
Following to the two-stage hypothesis of vehicle trajectory execution, this sub-module is dedicated to accurately developing the trajectory towards the predetermined goal. According to \cite{avidan_human_2022}, we developed a goal-based neural social force model. As we mentioned in \textbf{Section 2.1}, the state $q_i^t$ of the $i$th vehicle at any time $t$ can be observed on the highway. Then a sequence of vehicle state can be represented as a function of time $q(t)$. Similarly, the sequences of neighborhood vehicle states is also a function of time $\Omega(t)$. Therefore, the vehicle dynamics in a highway in this neural social force model can be formulated as follows:

\begin{equation}
\frac{dq}{dt}(t) = f_{\theta, \phi}(t, q(t), \Omega(t), q^T, E),
\end{equation}

\begin{linenomath*}
\vspace{1em}
\end{linenomath*}

\noindent where $\theta$ and $\phi$ are interpretable parameters presented later and uninterpretable parameters in neural networks. The vehicle dynamics determined by function $f$, which is governed by time $t$, the current state $q(t)$, its neighboring vehicle state $\Omega(t)$ and the environment $E$.

Given the initial and final condition $q(0) = q^0$ and $q(T) = q^T$, then we have the following representation:

\begin{equation}
q^T = q^0 + \int_{t=0}^{T} f_{\theta, \phi}(t, q(t), \Omega(t), q^T, E) dt
\end{equation}

\begin{linenomath*}
\vspace{1em}
\end{linenomath*}

Assuming $p(t)$ is second-order differentiable, we expands $q(t)$ using Taylor’s series for a first-order approximation:

\begin{equation}
q(t + \Delta t) \approx q(t) + \dot{q}(t) \Delta t = 
\begin{pmatrix}
p(t) \\
\dot{p}(t)
\end{pmatrix}
+ \Delta t 
\begin{pmatrix}
\dot{p}(t) \\
\ddot{p}(t)
\end{pmatrix}
\end{equation}

\begin{linenomath*}
\vspace{1em}
\end{linenomath*}

\noindent where $\Delta t$ is the time step. The stochasticity $\alpha(t, q^{t:t-M})$ is assumed to only influence $p$. \textbf{Equation 11} is general and any dynamical system with second-order differentiability can be employed here. 

Further, we hypothesize that each vehicle behaves as a particle in a particle system and follows Newton's second law of motion. The acceleration $\ddot{p}(t)$ that we establish depends primarily on two component forces: the Goal attraction force $F_{goal}$ and the inter-vehicle and environmental repulsion $F_{rep}$.

\begin{equation}
\ddot{p}(t) = F_{\text{goal}}(t, q^T, q^t) + F_{\text{rep}}(t, q^t, \Omega^t)
\end{equation}

\begin{linenomath*}
\vspace{1em}
\end{linenomath*}

Unlike the traditional social force model, some parameters in our model are obtained through neural network training. We assume that the goal or destination $P^T$ of each trajectory is given, although in practice, the goal $P^T$ needs to be learned or sampled during prediction. Therefore, we employ the Goal Prediction sub-module described in \textbf{Section 3.3} to sample $P^T$ for each trajectory. For trajectory prediction, the Goal Prediction sub-module is a pre-trained model. When new trajectory data is collected during testing, the Goal Prediction sub-module samples the goal $P^T$ for each trajectory, and the Trajectory Prediction sub-module predicts the future trajectories.

Given the current state and goal, we calculate \(F_{\text{goal}}\) using the goal network \(NN_{\phi_1}\) as described in \textbf{Equation 13} and \(F_{\text{rep}}\) using the repulsion network \(NN_{\phi_2}\) as described in \textbf{Equation 14}. The goal network first encodes the current state \(q^t\), which is then input into a Long Short-Term Memory (LSTM) network to capture the dynamics. After a linear transformation, the LSTM output is concatenated with the embedded goal \(p^T\). Finally, the key parameter \(\tau\) is computed using a Multi-Layer Perceptron (MLP). The architecture of the collision network is similar. Each agent \(q_j^t\) in the neighborhood \(\Omega_n^t\) is encoded and concatenated with the encoded target vehicle state \(q_n^t\). The parameter \(k_{nj}\) is then computed. Through these steps, the interpretable key parameters \(\tau\) and \(k_{nj}\) for \(F_{\text{goal}}\) and \(F_{\text{rep}}\) are derived.

\vspace{1em}
\noindent \textbf{Goal attraction:} A vehicle navigates towards its destination or goal driven by its intrinsic driving intentions, which we abstractly model as an attraction generated by the goal. At time \( t \), a vehicle has a desired driving direction \( e^t \), determined by the goal \( p^T \) and the current position \( p^t \): \( e^t = \frac{p^T - p^t}{\| p^T - p^t \|} \). Without any interfering forces, the vehicle adjusts its current velocity to align with the desired velocity \( v_{\text{des}}^t = v_0^t e^t \), where \( v_0^t \) and \( e^t \) denote the magnitude and direction of the velocity, respectively. Unlike the static \( v_0 \) used in previous models, we dynamically update \( v_0^t \) at each time step to reflect the changing desired speed as the vehicle approaches its destination: \( v_0^t = \frac{\| p^T - p^t \|}{(T-t) \Delta t} \). The goal attraction force \( F_{\text{goal}} \) indicates the vehicle's natural tendency to adjust its current velocity \( \dot{p}^t \) to the desired velocity \( v_{\text{des}}^t \) within time \( \tau \).

\begin{equation}
F_{\text{goal}} = \frac{1}{\tau} (v_{\text{des}}^t - \dot{p}^t) \quad \text{where} \quad \tau = NN_{\phi_1}(q^t, p^T)
\end{equation}

\begin{linenomath*}
\vspace{1em}
\end{linenomath*}

\noindent where $\tau$ is learned through a neural network (NN) parameterized by $\phi_1$.

\vspace{1em}
\noindent \textbf{Inter vehicle and environment repulsion:} Besides the goal attraction force, vehicles on a highway are subject to certain repulsive forces. The repulsive force stems from two major components: (1) maintaining a safety distance from surrounding vehicles to avoid collisions, and (2) adhering to lane constraints, where lane changes can be made at the dashed lines but never across the road boundary lines. Given a target vehicle \( n \), a neighboring vehicle \( j \in \Omega_{n}^t \) at relative position \( \mathbf{r}_{nj} = \mathbf{p}_n^t - \mathbf{p}_j^t \), and lane boundaries \( l \in \Lambda_{n}^t \) at distances \( d_{nl} \), the repulsive force \( \mathbf{F}_{\text{rep}}^{nj} \) exerted on vehicle \( n \) is based on the gradient of the total repulsive potential field \( U_{\text{total}} \):

\begin{equation}
\mathbf{F}_{\text{rep}}^{nj} = -\nabla_{\mathbf{r}_{nj}} U_{\text{total}}
\end{equation}

\begin{linenomath*}
\vspace{1em}
\end{linenomath*}

Under previous assumptions, the total repulsive potential field \( U_{\text{total}} \) consists of two components: the potential field from surrounding vehicles \( U_{\text{vehicles}} \) and the potential field from lane boundaries \( U_{\text{lines}} \). The total repulsive potential field is given by:

\begin{equation}
U_{\text{total}} = \sum_{j \in \Omega_{n}^t} \left( r_{\text{col}} k_{nj} e^{-\frac{\| \mathbf{r}_{nj} \|}{r_{\text{col}}}} \right) + \sum_{l \in \Lambda_{n}^t} U_{\text{line}, l}
\end{equation}

\begin{linenomath*}
\vspace{1em}
\end{linenomath*}

Here, \( r_{\text{col}} \) is a scaling factor for the repulsive potential, \( k_{nj} \) is a coefficient representing the interaction strength between the target vehicle \( n \) and the neighboring vehicle \( j \), and \( \| \mathbf{r}_{nj} \| \) is the distance between these vehicles. The term \( U_{\text{line}, l} \) represents the potential field contributions from the lane boundaries and includes the interaction strength parameter \( k_{nl} \), given by:

\begin{equation}
U_{\text{line}, l} = 
\begin{cases} 
k_{nl} e^{-(d_{nl})^2} & \text{for center lines} \\
k_{nl} \frac{0.5}{d_{nl}^2} & \text{for boundary lines}
\end{cases}
\end{equation}

\begin{linenomath*}
\vspace{1em}
\end{linenomath*}

The selection of \( U_{\text{line}, l} \) is based on the different characteristics and roles of center lines and boundary lines in the highway, as referenced in \cite{li_dynamic_2022, han_spatial-temporal_2023}. Center lines are typically crossable by vehicles, indicating vehicles can change lanes for overtaking or other maneuvers. Therefore, an exponential decay model \( k_{nl} e^{-(d_{nl})^2} \) is used, which provides a decent potential field when close to the center line but rapidly diminishes as the distance increases, reflecting the flexibility and frequent interaction with the center line. Conversely, boundary lines are generally non-crossable, serving as strict limits to the vehicle's movement, such as road edges or barriers. Thus, an inverse square model \( k_{nl} \frac{0.5}{d_{nl}^2} \) is used for boundary lines. This generates a stronger and more persistent repulsive force as the vehicle approaches the boundary line, ensuring that the vehicle maintains a safe distance and avoids collisions with these non-crossable boundaries. 

Moreover, to ensure that the total repulsive potential field \( U_{\text{total}} \) varies over time, we treat \( k_{nj} \) and \( k_{nl} \) as learnable dynamic variables. We define \([k_{nj}, k_{nl}] = a \ast \text{sigmoid}(NN_{\phi_2}(q_n^t, q_j^t, j \in \Omega_n^t, p_l^t, l \in \Lambda_n^t))\), and adjust the hyperparameter \( a \) to 
guarantee the learned \( k_{nj} \) and \( k_{nl} \) are realistic.

\section{3.Results and discussion}
\subsection{3.1 Experimental Dataset and Setting}
Our research employs two datasets. The first one, known as the Next Generation Simulation (NGSIM) dataset, provides comprehensive vehicle trajectory data from eastbound I-80 in the San Francisco Bay area and southbound US 101 in Los Angeles. This data, collected by the U.S. Department of Transportation in 2015, captures real-world highway conditions via overhead cameras operating at 10 Hz. The second dataset is HighD, derived from drone video recordings at 25 Hz between 2017 and 2018 around Cologne, Germany. It covers approximately 420 meters of two-way roads and documents 110,000 vehicles, including both cars and trucks, traveling a cumulative distance of 45,000 km. We trim each trajectory to a length of 8 seconds. Given 3 seconds of trajectory data, we train the model to predict the remaining 5 seconds of trajectory points. Specifically, for the NGSIM dataset, 30 frames are used to predict 50 future frames, while for the HighD dataset, 75 frames are input to predict 125 future frames.

\subsection{3.2 Metrics}
We evaluate our proposed and compared methods using three metrics: Root Mean Square Error (RMSE), Average Displacement Error (ADE), and Final Displacement Error (FDE). Given a true future trajectory (ground truth) \(\{p_t = (x_t, y_t)\}_{t=T_{\text{obs}}+1}^T\) and the corresponding predicted trajectory \(\{\hat{p}_t = (\hat{x}_t, \hat{y}_t)\}_{t=T_{\text{obs}}+1}^T\), these metrics measure the \( \ell_2 \) distance between the ground truth and the predicted trajectory. ADE calculates the average \( \ell_2 \) distance between the predicted trajectory and the ground truth trajectory across the entire prediction period. FDE calculates the \( \ell_2 \) distance between the predicted final position and the actual final position at the end of the prediction period. RMSE assesses overall accuracy by computing the square root of the average squared differences between predicted and actual positions across all time steps. The metrics are defined as follows:

\begin{equation}
\begin{aligned}
    \text{ADE} &= \frac{1}{T_{\text{pred}}} \sum_{t=T_{\text{obs}}+1}^{T} \sqrt{(x_t - \hat{x}_t)^2 + (y_t - \hat{y}_t)^2}, \\
    \text{FDE} &= \sqrt{(x_T - \hat{x}_T)^2 + (y_T - \hat{y}_T)^2}, \\
    \text{RMSE} &= \sqrt{\frac{1}{T - T_{\text{obs}}} \sum_{t=T_{\text{obs}}+1}^{T} ((x_t - \hat{x}_t)^2 + (y_t - \hat{y}_t)^2)}
\end{aligned}
\end{equation}

\begin{linenomath*}
\vspace{1em}
\end{linenomath*}

\subsection{3.3 Baseline models}
We compare our model with the following baselines:

\begin{itemize}
    \item \textbf{Social-LSTM (S-LSTM) \cite{alahi2016social}:} This model uses a shared LSTM to encode the raw trajectory data for each vehicle. The extracted features from different vehicles are then aggregated using a social pooling layer.
    
    \item \textbf{Convolutional Social-LSTM (CS-LSTM) \cite{deo_convolutional_2018}:} Unlike S-LSTM, this model captures social interactions by stacking convolutional and pooling layers, and it considers multi-modality based on the predicted intention.
    
    \item \textbf{Planning-informed Prediction (PiP) \cite{song2020pip}:} This model integrates trajectory prediction with the planning of the target vehicle by conditioning on multiple candidate trajectories for the target vehicle.
    
    \item \textbf{Spatial-temporal dynamic attention network (STDAN) \cite{chen_intention-aware_2022}:} This paper introduces a spatiotemporal dynamic attention network designed for intention-aware vehicle trajectory prediction. It uses hierarchical modules to capture various levels of social and temporal features, employing a multi-head attention mechanism to extract data from raw trajectories and a novel feature fusion method for joint intention recognition and trajectory prediction. 

    \item \textbf{GNP:} Our proposed model integrates physics and deep learning for goal-based trajectory prediction in this paper.
\end{itemize}

\subsection{3.4 Quantitative Results}
Table 1 shows the comparison results with the baseline models. Our proposed models consistently outperform S-LSTM, CS-LSTM, PIP, and STDAN models in all 1-5 second predictions. We use the RMSE metric to compare the accuracy of the models. Our proposed model is evaluated with three metrics: ADE, FDE, and RMSE. The results with the best accuracy are highlighted in bold. The goal-based framework achieves excellent results in terms of accuracy, effectively capturing the goals even though the full trajectory prediction is derived from a physical model. Additionally, the model performs better on the HighD data compared to NGSIM data. This may be because the scenarios in HighD data have fewer lanes and the shorter distances required for lane changes, which make the prediction task less challenging.

\begin{table}[!ht]
    \caption{Prediction Error Obtained by Different Models}
    \label{tab:rmse}
    \centering
    \begin{tabular}{lccccc|c}
        \toprule
        Dataset & Horizon & \multicolumn{4}{c}{RMSE} & \multicolumn{1}{c}{ADE/FDE/RMSE} \\
        & (s) & S-LSTM & CS-LSTM & PiP & STDAN & GNP \\
        \midrule
        NGSIM & 10 & 0.65 & 0.61 & 0.55 & 0.42 & 0.3387/0.6228/ \textbf{0.2733} \\
        & 20 & 1.31 & 1.27 & 1.18 & 1.01 & 0.67/1.30/ \textbf{0.55} \\
        & 30 & 2.16 & 2.08 & 1.94 & 1.69 & 1.03/2.07/ \textbf{0.86} \\
        & 40 & 3.25 & 3.10 & 2.88 & 2.56 & 1.43/2.93/ \textbf{1.21} \\
        & 50 & 4.55 & 4.37 & 4.04 & 3.67 & 1.86/3.91/ \textbf{1.59} \\
        \midrule
        HighD & 10 & 0.22 & 0.22 & 0.17 & 0.29 & 0.11/0.19/ \textbf{0.09} \\
        & 20 & 0.62 & 0.61 & 0.50 & 0.68 & 0.21/0.38/ \textbf{0.17} \\
        & 30 & 1.27 & 1.24 & 1.05 & 1.17 & 0.32/0.57/ \textbf{0.26} \\
        & 40 & 2.15 & 2.10 & 1.76 & 1.88 & 0.44/0.80/ \textbf{0.37} \\
        & 50 & 3.41 & 3.27 & 2.63 & 2.76 & 0.59/1.07/ \textbf{0.50} \\
        \bottomrule
    \end{tabular}
\end{table}

\subsection{3.5 Visualization Results}
\begin{figure}[!ht]
  \centering
  \includegraphics[width=1\textwidth]{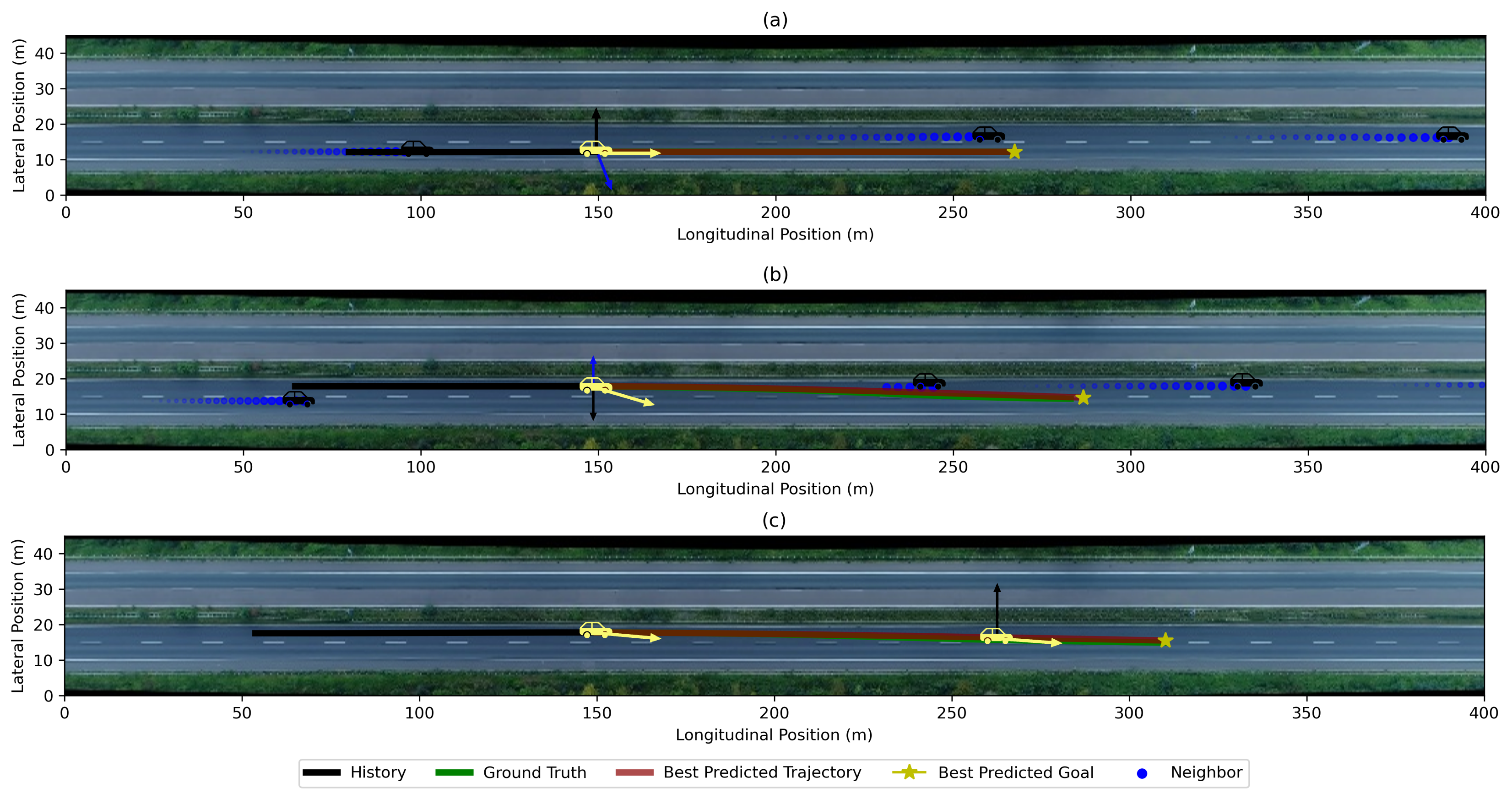}
  \caption{Interpretability of vehicle trajectory predictions visualized in three example scenarios.Yellow arrow denotes the goal attraction force, blue arrow denotes the combined repulsive forces generated by the neighboring vehicles and the black arrow indicates the combined repulsive forces exerted by the lane lines.}\label{fig interpretability}
\end{figure}

Unlike deep learning-based trajectory prediction models, our GNP model inherently provides interpretable results by analyzing the forces acting on the target vehicle. Figure~\ref{fig interpretability} illustrates several examples of the GNP model applied to a two-lane scenario in the HighD dataset, viewed from a right-to-left vehicle travel perspective for clarity. In Figure~\ref{fig interpretability} (a), the vehicle continues traveling straight in the right lane. The yellow car represents the current position of the target vehicle. The GOAL attraction force (yellow arrow) pulls it forward due to the two cars ahead in the left lane and a clear path in the right lane. The vehicle is repelled to the right by neighboring vehicles (blue arrows) and to the left by the lane boundary (black arrows), balancing the repulsion forces to keep it in the right lane. In Figure~\ref{fig interpretability} (b), the opposite scenario unfolds. Here, the target vehicle faces two cars directly ahead in its current lane, while the adjacent right lane is clear. This prompts the vehicle to initiate a lane change to the right. 

Figure~\ref{fig interpretability} (c) depicts a more straightforward case. The vehicle is not influenced by any neighboring vehicle and is centered in its lane, thus unaffected by repulsion forces. Driven solely by the goal's attraction force, it changes lanes to the right. As it nears the centerline, a leftward repulsion force emerges but does not prevent the lane change due to the dominant goal attraction force. Subsequent ablation experiments demonstrate that the attraction force predominantly influences the model's prediction accuracy.

\begin{figure}[!ht]
  \centering
  \includegraphics[width=0.6\textwidth]{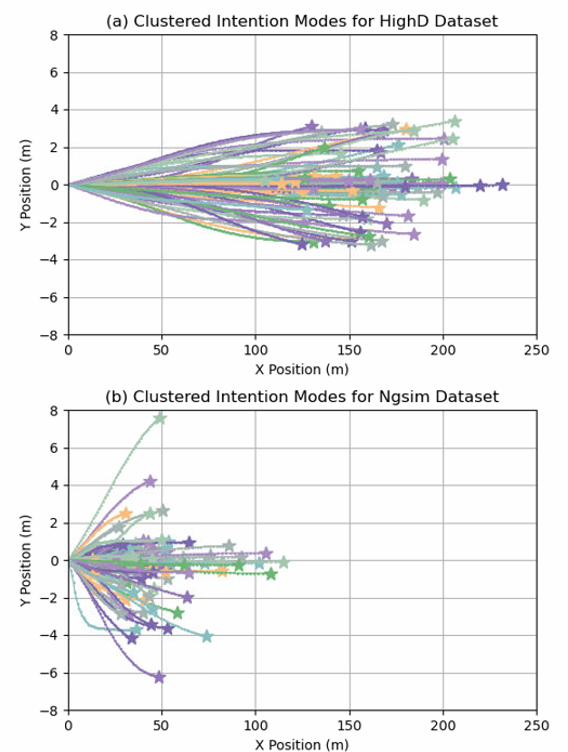}
  \caption{Clustered intention modes from HighD (a) and Ngsim (b) dataset}\label{fig intention mode}
\end{figure}

Figure~\ref{fig intention mode} illustrates the intention modes mentioned in \textbf{Section 2.3}, derived from clustering the HighD and Ngsim datasets. In Figure~\ref{fig intention mode} (a), three primary trends are identified: straight forward, left lane change, and right lane change, with straight travel being the most common. For instance, in left lane changes, an upward-opening curve indicates a gradual decrease in the trend to the left, suggesting the vehicle is completing a lane change. Conversely, a downward-opening curve indicates a faster trend to the left, suggesting the vehicle is beginning to change lanes. Figure~\ref{fig intention mode} (b) reveals that, unlike HighD data, Ngsim data shows a lower traveling speed but a wider range of lane changes, due to differences in traffic volume and number of lanes. However, both datasets display the same trends of straight travel, left lane change, and right lane change. Therefore, the intention modes effectively capture common motion behaviors and underlying driving intentions. We will later demonstrate the impact of these intention modes on prediction accuracy through ablation experiments.

\begin{figure}[!ht]
  \centering
  \includegraphics[width=1.0\textwidth]{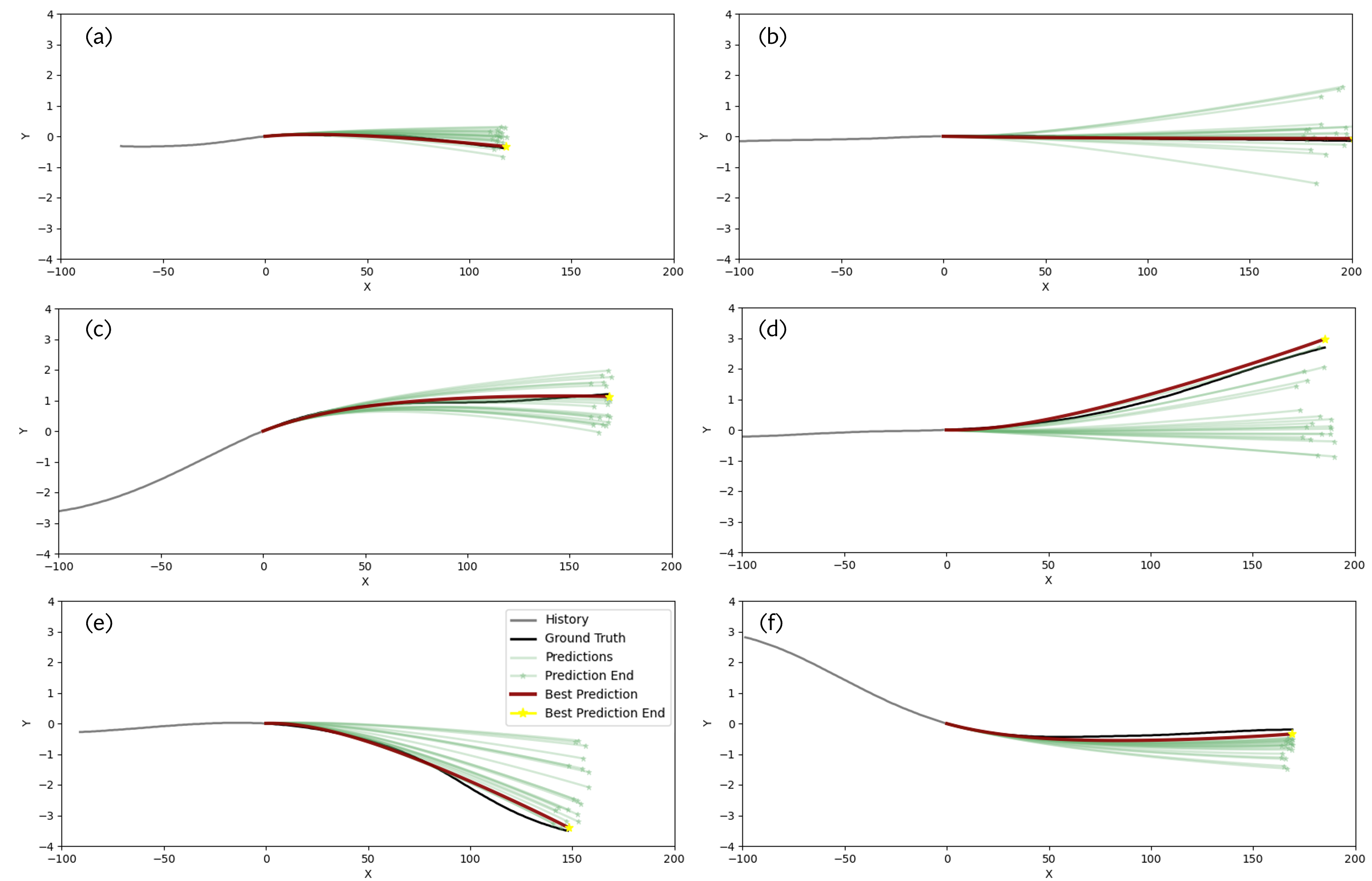}
  \caption{Multiple prediction results on 3 different bahaviors: straight ahead (a, b), left lane change (c, d), and right lane change (e,f) }\label{fig multiple predictions}
\end{figure}

Figure~\ref{fig multiple predictions} demonstrates the GNP model's robust prediction performance across different driving behaviors, including straight ahead (a, b), left lane change (c, d), and right lane change (e, f). The best predicted trajectory is highlighted in dark red, while lime green represents other potential predictions. In Figure~\ref{fig multiple predictions} (a) and (b), the model accurately predicts straight-ahead trajectories at different speeds, while also considering the potential for lane changes. Figure~\ref{fig multiple predictions} (c) shows a historical trajectory with a clear left lane-change tendency, leading to more concentrated predictions. In contrast, Figure~\ref{fig multiple predictions} (d) has a nearly straight historical trajectory, but the model still accounts for a possible left lane change. The right lane change predictions in Figure~\ref{fig multiple predictions} (e) and (f) exhibit similar characteristics to the left lane change predictions. Thus, the GNP effectively models multiple goals and ultimately capturing the vehicle's possible future motion behaviors with high accuracy.

\subsection{3.6 Ablation experiments}
We evaluate the significance of GNP's key components by comparing various model variants on 5-second predictions using the HighD dataset. In Table~\ref{ablation}, "IM" denotes the Intention Modes derived from clustering in the Goal Prediction module, "F\_Goal" represents the GOAL attraction force in the Trajectory Prediction module, and "F\_Rep" indicates the repulsive force generated by neighboring vehicles and lane lines. From Table~\ref{ablation}, we observe that variant 1, which excludes IM, cannot accurately predict the goal due to the absence of intention modeling, leading to inaccurate prediction results. Furthermore, variants 3 and 4 demonstrate that F\_Goal has already achieved great performance even without the repulsion force. This can be attributed to two factors. Firstly, the Trajectory Prediction module is trained progressively, with F\_Goal capturing the majority of features initially, while F\_Rep provides additional details. Secondly, highway driving behaviors are typically straightforward; vehicles generally achieve their intentions of either maintaining straight paths or changing lanes, with surrounding vehicles and lane lines serving mainly as constraints to avoid collisions. Consequently, the attraction force exerts a more significant influence on the vehicle's trajectory.

\begin{table}[!ht]
    \caption{Comparison of Different Variants in Ablation Experiments}
    \label{ablation}
    \centering
    \begin{tabular}{lcccccc}
        \toprule
        Variant & IM & F\_GOAL & F\_REP & & ADE/FDE/RMSE \\
        \midrule
        (1) & \ding{55} & \ding{51} & \ding{51} & & 2.21/4.02/3.53 \\
        (2) & \ding{55} & \ding{51} & \ding{55} & & 2.63/4.11/3.68 \\
        (3) & \ding{51} & \ding{51} & \ding{55} & & 0.87/1.46/0.77 \\
        (4) & \ding{51} & \ding{51} & \ding{51} & & \textbf{0.59/1.07/0.50} \\
        \bottomrule
    \end{tabular}
\end{table}




\section{4. Conclusion}
In this paper, we have presented a novel goal-based vehicle trajectory prediction model that integrates the strength of deep learning with a physics-based social force model for better accuracy and interpretability. By modeling both vehicle intention estimation and full trajectory prediction, our dual sub-modules model effectively addresses the inherent uncertainties in vehicle trajectory prediction. First, the Goal Prediction sub-module introduced in \textbf{Section 2.3} employs a multi-head attention mechanism to accurately capture driving intentions (goals), tackling the challenge of anticipating vehicle maneuvers over a specific period. Second, the Goal-based Neural Social Force Trajectory Prediction sub-module introduced in \textbf{Section 2.4} combines the interpretability of the social force model with the strong data-fitting capabilities of deep learning, progressively predicting the complete trajectory based on the estimated goals.

Our experimental results on two highway datasets demonstrate that our model achieves state-of-the-art long-term prediction accuracy, outperforming baseline models. Additionally, the integration of neural differential equations within the social force framework enhances the model's interpretability, providing transparent and trustworthy predictions through visualization. By effectively balancing accuracy and interpretability, our model offers a promising solution for addressing trajectory prediction challenges in intelligent transportation systems and autonomous driving.

In the future, we plan to improve the clustered intention modes with more advanced designs to better explore the potential driving goals of vehicles and estimate the goals more accurately. Furthermore, the potential field model used to calculate the repulsion force in this paper is relatively basic; a more refined potential field can be developed to better capture the inter-vehicle and vehicle-road relationships. Finally, our experiments were limited to straight highway sections, so future work should consider more complex scenarios such as urban road networks and highways with ramps.

\section{Acknowledgements}
We acknowledge the support provided by the Traffic Operations and Safety (TOPS) Laboratory at the University of Wisconsin-Madison. The ideas and views expressed in this paper are those of the authors and do not necessarily reflect the view of the University of Wisconsin-Madison.

\section{Author Contributions}
The authors confirm contribution to the paper as follows: conceptualization, methodology, validation, writing–original draft: Rui GAN, Bocheng AN, Haotian Shi, Pei Li; software, formal analysis: Rui GAN, Pei Li, Junyi Ma; review \& editing: Haotian Shi, Pei Li, Chengyuan Ma, and Bin RAN. All authors reviewed the results and approved the final version of the manuscript.

\newpage

\bibliographystyle{trb}
\bibliography{2024trb_GNP}

\begin{thebibliography}{31}
\providecommand{\natexlab}[1]{#1}

\bibitem[{Shi et~al.(2023{\natexlab{a}})Shi, Chen, Zheng, Wang, Zhou, and Ran}]{shi2023deep}
Shi, H., D.~Chen, N.~Zheng, X.~Wang, Y.~Zhou, and B.~Ran, A deep reinforcement learning based distributed control strategy for connected automated vehicles in mixed traffic platoon. \emph{Transportation Research Part C: Emerging Technologies}, Vol. 148, 2023{\natexlab{a}}, p. 104019.

\bibitem[{Lefèvre et~al.(2014)Lefèvre, Vasquez, and Laugier}]{lefevre_survey_2014}
Lefèvre, S., D.~Vasquez, and C.~Laugier, A survey on motion prediction and risk assessment for intelligent vehicles. \emph{ROBOMECH Journal}, Vol.~1, No.~1, 2014, p.~1.

\bibitem[{Goli et~al.(2018)Goli, Far, and Fapojuwo}]{goli2018vehicle}
Goli, S.~A., B.~H. Far, and A.~O. Fapojuwo, Vehicle Trajectory Prediction with Gaussian Process Regression in Connected Vehicle Environment $\star$. In \emph{2018 IEEE Intelligent Vehicles Symposium (IV)}, IEEE, 2018, pp. 550--555.

\bibitem[{Altch{\'e} and de~La~Fortelle(2017)}]{altche2017lstm}
Altch{\'e}, F. and A.~de~La~Fortelle, An LSTM network for highway trajectory prediction. In \emph{2017 IEEE 20th international conference on intelligent transportation systems (ITSC)}, IEEE, 2017, pp. 353--359.

\bibitem[{Benterki et~al.(2019)Benterki, Judalet, Choubeila, and Boukhnifer}]{benterki2019long}
Benterki, A., V.~Judalet, M.~Choubeila, and M.~Boukhnifer, Long-term prediction of vehicle trajectory using recurrent neural networks. In \emph{IECON 2019-45th Annual Conference of the IEEE Industrial Electronics Society}, IEEE, 2019, Vol.~1, pp. 3817--3822.

\bibitem[{Park et~al.(2018)Park, Kim, Kang, Chung, and Choi}]{park2018sequence}
Park, S.~H., B.~Kim, C.~M. Kang, C.~C. Chung, and J.~W. Choi, Sequence-to-sequence prediction of vehicle trajectory via LSTM encoder-decoder architecture. In \emph{2018 IEEE intelligent vehicles symposium (IV)}, IEEE, 2018, pp. 1672--1678.

\bibitem[{Chen et~al.(2022{\natexlab{a}})Chen, Zhang, Zhao, Cai, Wang, and Ye}]{chen2022vehicle}
Chen, X., H.~Zhang, F.~Zhao, Y.~Cai, H.~Wang, and Q.~Ye, Vehicle trajectory prediction based on intention-aware non-autoregressive transformer with multi-attention learning for Internet of Vehicles. \emph{IEEE Transactions on Instrumentation and Measurement}, Vol.~71, 2022{\natexlab{a}}, pp. 1--12.

\bibitem[{Zhang et~al.(2022)Zhang, Feng, Wu, and He}]{zhang2022trajectory}
Zhang, K., X.~Feng, L.~Wu, and Z.~He, Trajectory prediction for autonomous driving using spatial-temporal graph attention transformer. \emph{IEEE Transactions on Intelligent Transportation Systems}, Vol.~23, No.~11, 2022, pp. 22343--22353.

\bibitem[{Deo and Trivedi(2018)}]{deo_convolutional_2018}
Deo, N. and M.~M. Trivedi, Convolutional {Social} {Pooling} for {Vehicle} {Trajectory} {Prediction}. In \emph{2018 {IEEE}/{CVF} {Conference} on {Computer} {Vision} and {Pattern} {Recognition} {Workshops} ({CVPRW})}, IEEE, Salt Lake City, UT, USA, 2018, pp. 1549--15498.

\bibitem[{Messaoud et~al.(2019)Messaoud, Yahiaoui, Verroust-Blondet, and Nashashibi}]{messaoud2019non}
Messaoud, K., I.~Yahiaoui, A.~Verroust-Blondet, and F.~Nashashibi, Non-local social pooling for vehicle trajectory prediction. In \emph{2019 IEEE Intelligent Vehicles Symposium (IV)}, IEEE, 2019, pp. 975--980.

\bibitem[{Wang et~al.(2020)Wang, Zhao, Zhang, Cheng, and Yang}]{wang2020multi}
Wang, Y., S.~Zhao, R.~Zhang, X.~Cheng, and L.~Yang, Multi-vehicle collaborative learning for trajectory prediction with spatio-temporal tensor fusion. \emph{IEEE Transactions on Intelligent Transportation Systems}, Vol.~23, No.~1, 2020, pp. 236--248.

\bibitem[{Wu et~al.(2023)Wu, Zhou, Shi, Li, and Ran}]{wu_graph-based_2023}
Wu, K., Y.~Zhou, H.~Shi, X.~Li, and B.~Ran, \emph{Graph-{Based} {Interaction}-{Aware} {Multimodal} {2D} {Vehicle} {Trajectory} {Prediction} using {Diffusion} {Graph} {Convolutional} {Networks}}, 2023, arXiv:2309.01981 [cs].

\bibitem[{Li et~al.(2019)Li, Ying, and Chuah}]{li2019grip}
Li, X., X.~Ying, and M.~C. Chuah, Grip: Graph-based interaction-aware trajectory prediction. In \emph{2019 IEEE Intelligent Transportation Systems Conference (ITSC)}, IEEE, 2019, pp. 3960--3966.

\bibitem[{Feng et~al.(2019)Feng, Cen, Hu, and Zhang}]{feng2019vehicle}
Feng, X., Z.~Cen, J.~Hu, and Y.~Zhang, Vehicle trajectory prediction using intention-based conditional variational autoencoder. In \emph{2019 IEEE Intelligent Transportation Systems Conference (ITSC)}, IEEE, 2019, pp. 3514--3519.

\bibitem[{Gupta et~al.(2018)Gupta, Johnson, Fei-Fei, Savarese, and Alahi}]{gupta2018social}
Gupta, A., J.~Johnson, L.~Fei-Fei, S.~Savarese, and A.~Alahi, Social gan: Socially acceptable trajectories with generative adversarial networks. In \emph{Proceedings of the IEEE conference on computer vision and pattern recognition}, 2018, pp. 2255--2264.

\bibitem[{Ghoul et~al.(2022)Ghoul, Messaoud, Yahiaoui, Verroust-Blondet, and Nashashibi}]{ghoul_lightweight_2022}
Ghoul, A., K.~Messaoud, I.~Yahiaoui, A.~Verroust-Blondet, and F.~Nashashibi, A {Lightweight} {Goal}-{Based} model for {Trajectory} {Prediction}. In \emph{2022 {IEEE} 25th {International} {Conference} on {Intelligent} {Transportation} {Systems} ({ITSC})}, IEEE, Macau, China, 2022, pp. 4209--4214.

\bibitem[{Tran et~al.(2021)Tran, Le, and Tran}]{tran_goal-driven_2021}
Tran, H., V.~Le, and T.~Tran, Goal-driven {Long}-{Term} {Trajectory} {Prediction}. In \emph{2021 {IEEE} {Winter} {Conference} on {Applications} of {Computer} {Vision} ({WACV})}, IEEE, Waikoloa, HI, USA, 2021, pp. 796--805.

\bibitem[{Ghoul et~al.(2023)Ghoul, Yahiaoui, Verroust-Blondet, and Nashashibi}]{ghoul_interpretable_2023}
Ghoul, A., I.~Yahiaoui, A.~Verroust-Blondet, and F.~Nashashibi, Interpretable {Goal}-{Based} model for {Vehicle} {Trajectory} {Prediction} in {Interactive} {Scenarios}. In \emph{2023 {IEEE} {Intelligent} {Vehicles} {Symposium} ({IV})}, IEEE, Anchorage, AK, USA, 2023, pp. 1--6.

\bibitem[{Zhao et~al.(2021)Zhao, Gao, Lan, Sun, Sapp, Varadarajan, Shen, Shen, Chai, Schmid et~al.}]{zhao2021tnt}
Zhao, H., J.~Gao, T.~Lan, C.~Sun, B.~Sapp, B.~Varadarajan, Y.~Shen, Y.~Shen, Y.~Chai, C.~Schmid, et~al., Tnt: Target-driven trajectory prediction. In \emph{Conference on Robot Learning}, PMLR, 2021, pp. 895--904.

\bibitem[{Gu et~al.(2021)Gu, Sun, and Zhao}]{gu_densetnt_2021}
Gu, J., C.~Sun, and H.~Zhao, {DenseTNT}: {End}-to-end {Trajectory} {Prediction} from {Dense} {Goal} {Sets}. In \emph{2021 {IEEE}/{CVF} {International} {Conference} on {Computer} {Vision} ({ICCV})}, IEEE, Montreal, QC, Canada, 2021, pp. 15283--15292.

\bibitem[{Mangalam et~al.(2020)Mangalam, Girase, Agarwal, Lee, Adeli, Malik, and Gaidon}]{mangalam_it_2020}
Mangalam, K., H.~Girase, S.~Agarwal, K.-H. Lee, E.~Adeli, J.~Malik, and A.~Gaidon, \emph{It {Is} {Not} the {Journey} but the {Destination}: {Endpoint} {Conditioned} {Trajectory} {Prediction}}, 2020, arXiv:2004.02025 [cs].

\bibitem[{Yue et~al.(2022)Yue, Manocha, and Wang}]{avidan_human_2022}
Yue, J., D.~Manocha, and H.~Wang, Human {Trajectory} {Prediction} via {Neural} {Social} {Physics}. In \emph{Computer {Vision} – {ECCV} 2022} (S.~Avidan, G.~Brostow, M.~Cissé, G.~M. Farinella, and T.~Hassner, eds.), Springer Nature Switzerland, Cham, Vol. 13694, 2022, pp. 376--394, series Title: Lecture Notes in Computer Science.

\bibitem[{Geng et~al.(2023)Geng, Li, Xia, and Chen}]{geng_physics-informed_2023}
Geng, M., J.~Li, Y.~Xia, and X.~M. Chen, A physics-informed {Transformer} model for vehicle trajectory prediction on highways. \emph{Transportation Research Part C: Emerging Technologies}, Vol. 154, 2023, p. 104272.

\bibitem[{Huang et~al.(2011)Huang, Fellendorf, and Sch{\"o}nauer}]{huang2011social}
Huang, W., M.~Fellendorf, and R.~Sch{\"o}nauer, Social force based vehicle model for 2-dimensional spaces. In \emph{91st Annual Meeting of the Transportation Research Board. Washington, DC, USA}, 2011.

\bibitem[{Shi et~al.(2023{\natexlab{b}})Shi, Wang, Zhou, and Hua}]{shi_trajectory_2023}
Shi, L., L.~Wang, S.~Zhou, and G.~Hua, Trajectory {Unified} {Transformer} for {Pedestrian} {Trajectory} {Prediction}. In \emph{2023 {IEEE}/{CVF} {International} {Conference} on {Computer} {Vision} ({ICCV})}, IEEE, Paris, France, 2023{\natexlab{b}}, pp. 9641--9650.

\bibitem[{He et~al.(2016)He, Zhang, Ren, and Sun}]{he2016deep}
He, K., X.~Zhang, S.~Ren, and J.~Sun, Deep residual learning for image recognition. In \emph{Proceedings of the IEEE conference on computer vision and pattern recognition}, 2016, pp. 770--778.

\bibitem[{Li et~al.(2022)Li, Gan, Ji, Qu, and Ran}]{li_dynamic_2022}
Li, L., J.~Gan, X.~Ji, X.~Qu, and B.~Ran, Dynamic {Driving} {Risk} {Potential} {Field} {Model} {Under} the {Connected} and {Automated} {Vehicles} {Environment} and {Its} {Application} in {Car}-{Following} {Modeling}. \emph{IEEE Transactions on Intelligent Transportation Systems}, Vol.~23, No.~1, 2022, pp. 122--141.

\bibitem[{Han et~al.(2023)Han, Zhao, Zhu, and Song}]{han_spatial-temporal_2023}
Han, J., J.~Zhao, B.~Zhu, and D.~Song, Spatial-{Temporal} {Risk} {Field} for {Intelligent} {Connected} {Vehicle} in {Dynamic} {Traffic} and {Application} in {Trajectory} {Planning}. \emph{IEEE Transactions on Intelligent Transportation Systems}, Vol.~24, No.~3, 2023, pp. 2963--2975.

\bibitem[{Alahi et~al.(2016)Alahi, Goel, Ramanathan, Robicquet, Fei-Fei, and Savarese}]{alahi2016social}
Alahi, A., K.~Goel, V.~Ramanathan, A.~Robicquet, L.~Fei-Fei, and S.~Savarese, Social lstm: Human trajectory prediction in crowded spaces. In \emph{Proceedings of the IEEE conference on computer vision and pattern recognition}, 2016, pp. 961--971.

\bibitem[{Song et~al.(2020)Song, Ding, Chen, Shen, Wang, and Chen}]{song2020pip}
Song, H., W.~Ding, Y.~Chen, S.~Shen, M.~Y. Wang, and Q.~Chen, Pip: Planning-informed trajectory prediction for autonomous driving. In \emph{Computer Vision--ECCV 2020: 16th European Conference, Glasgow, UK, August 23--28, 2020, Proceedings, Part XXI 16}, Springer, 2020, pp. 598--614.

\bibitem[{Chen et~al.(2022{\natexlab{b}})Chen, Zhang, Zhao, Hu, Tan, and Yang}]{chen_intention-aware_2022}
Chen, X., H.~Zhang, F.~Zhao, Y.~Hu, C.~Tan, and J.~Yang, Intention-{Aware} {Vehicle} {Trajectory} {Prediction} {Based} on {Spatial}-{Temporal} {Dynamic} {Attention} {Network} for {Internet} of {Vehicles}. \emph{IEEE Transactions on Intelligent Transportation Systems}, Vol.~23, No.~10, 2022{\natexlab{b}}, pp. 19471--19483.

\end{thebibliography}
\end{document}